\pdfoutput=1
\documentclass[10pt,twocolumn,letterpaper]{article}

\usepackage{iccv}
\usepackage{times}
\usepackage{epsfig}
\usepackage{graphicx}
\usepackage{amsmath}
\usepackage{amssymb}
\usepackage{color}
\usepackage{algorithm}
\usepackage{algorithmic}
\usepackage{caption}
\usepackage[table]{xcolor}
\usepackage{epstopdf}
\usepackage{enumitem}
\DeclareGraphicsExtensions{.pdf,.png,.jpg}

\newcommand{\figref}[1]{Fig.~\ref{#1}}

\newcommand{\tabfref}[1]{Table~\ref{#1}}

\newcommand{\eqnref}[1]{Eq.~(\ref{#1})}
\newcommand{\secref}[1]{Sec.~\ref{#1}}

\usepackage[pagebackref=true,breaklinks=true,colorlinks,bookmarks=false]{hyperref}

\iccvfinalcopy 

\def\iccvPaperID{1306} 

\ificcvfinal\fi

\begin{document}

\title{Two-Phase Learning for Weakly Supervised Object Localization}

\author{\hspace{-0.1in}Dahun Kim\\
\hspace{-0.1in}KAIST\\
\hspace{-0.1in}{\tt\small mcahny@kaist.ac.kr}
\and
\hspace{-0.1in}Donghyeon Cho\\
\hspace{-0.1in}KAIST\\
\hspace{-0.1in}{\tt\small cdh12242@gmail.com}
\and 
\hspace{-0.1in}Donggeun Yoo\thanks{This work was done when he was in KAIST. He is currently working in Lunit Inc.}\\
\hspace{-0.1in}KAIST\\
\hspace{-0.1in}{\tt\small dgyoo@rcv.kaist.ac.kr}
\and
In So Kweon\\
KAIST\\
{\tt\small iskweon@kaist.ac.kr}
}

\maketitle

\begin{abstract}
Weakly supervised semantic segmentation and localization have a problem of focusing only on the most important parts of an image since they use only image-level annotations. In this paper, we solve this problem fundamentally via two-phase learning. Our networks are trained in two steps. In the first step, a conventional fully convolutional network (FCN) is trained to find the most discriminative parts of an image. In the second step, the activations on the most salient parts are suppressed by inference conditional feedback, and then the second learning is performed to find the area of the next most important parts. By combining the activations of both phases, the entire portion of the target object can be captured. Our proposed training scheme is novel and can be utilized in well-designed techniques for weakly supervised semantic segmentation, salient region detection, and object location prediction. Detailed experiments demonstrate the effectiveness of our two-phase learning in each task. 
\end{abstract}

\section{Introduction}

The most fundamental task for image understanding is to localize objects in a scene where each object has different locations and scales. It provides clues to challenging vision problems such as object detection and semantic segmentation. In recent years, deep learning based methods~\cite{GirshickDDM14,girshick15ICCV15,Shaoqing15NIPS,Long15CVPR,Noh15ICCV,chen14deeplab,LongSD15fcn,XiaDDCY13bbox} have achieved remarkably improved performance for those tasks by virtue of a large amount of annotated data and GPU parallel processing. However, it is expensive and laborious to obtain huge amounts of annotations such as bounding boxes and pixel-level labels. Therefore, weakly supervised learning~\cite{Oquab15,zhou2016cvpr,Bilen16,KantorovOCL16,Pinheiro2015CVPR,kolesnikov2016seed,KolesnikovL16a,Papandreou_2015_ICCV,pathakICCV15ccnn,SalehASPGA16builtin,Wei15stc} using only image-level annotations has begun to attract attention and shown interesting results.

However, there is still a large gap between the object localization power of weakly supervised methods and that of fully supervised methods. One major reason is that the localizability of weakly supervised FCNs is inherently limited to finding the most discriminative parts, rather than estimating the complete extent of objects. This is because image-level annotations simply lack information on the spatial extent of objects. Most existing weakly supervised methods for object localization~\cite{Oquab15,zhou2016cvpr,ZhangLBSS16exci,CinbisVS17multifold}, detection~\cite{Bilen16,KantorovOCL16,CholakkalJR16scspm,ShiF16size,BilenPT15wdet,WangHRZM15tipdet}, and semantic segmentation~\cite{Pinheiro2015CVPR,kolesnikov2016seed,KolesnikovL16a,Papandreou_2015_ICCV,pathakICCV15ccnn,SalehASPGA16builtin,Wei15stc,VasconcelosVC06,VezhnevetsB10,XuSU14tellme} suffer from this chronic problem.

\begin{figure}[t]
\def\arraystretch{0.5}
\begin{tabular}{@{}c@{\hskip 0.01\linewidth}c@{\hskip 0.01\linewidth}c@{\hskip 0.01\linewidth}}

\includegraphics[width=0.325\linewidth]{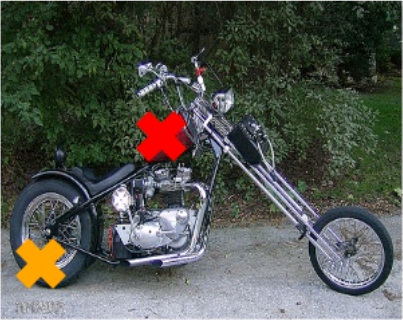} &
\includegraphics[width=0.325\linewidth]{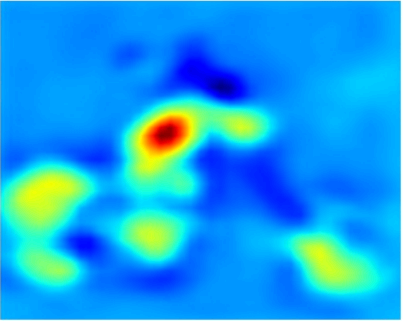} &
\includegraphics[width=0.325\linewidth]{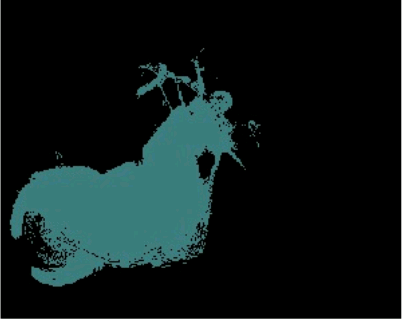}\\
{\small (a)} & {\small (b) } & {\small (c) }\\
\includegraphics[width=0.325\linewidth]{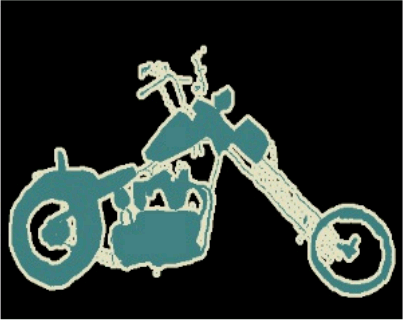} &
\includegraphics[width=0.325\linewidth]{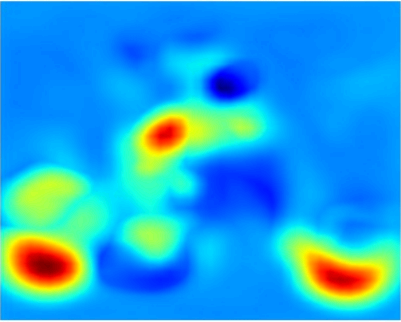} &
\includegraphics[width=0.325\linewidth]{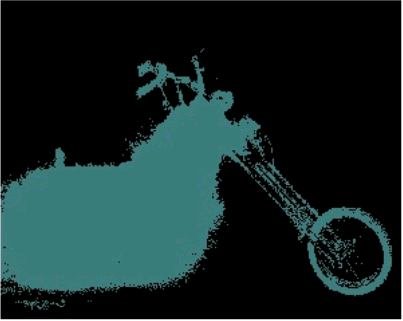} \\
{\small (d)} & {\small (e)} & {\small (f)} 
\end{tabular}
\caption{The effects of two-phase learning. (a) An input image, and estimated locations as the most (red) and the next (orange) important parts. (b) The heat map from the first network~\cite{zhou2016cvpr}. (c) The segmentation prediction of our baseline~\cite{kolesnikov2016seed}. (d) Ground truth segmentation mask. (e) The heat map from the proposed method. (f) The segmentation prediction using the proposed method.}
\label{fig:teaser}
\end{figure}

In this work, we overcome this problem fundamentally via two-phase learning. Our networks are trained in two phases. During the first phase, a conventional FCN is trained for image-level classification. At this time, pixels belonging to the most important parts in an image are revealed in a heat map, as shown in~\figref{fig:teaser}-(b).  During the second phase, another FCN is trained but the activations on highlighted regions in the first stage are suppressed via \textit{inference conditional feedback}. 
The underlying insight is that when the network is encouraged to discriminate images into their categories without knowledge of the most distinctive regions, the network will discover the next most discriminative parts of objects. At the inference stage, the entire portion of objects can be captured by combining activations of both phases, as illustrated in~\figref{fig:teaser}-(e). In other words, two-phase learning solves the fundamental problem that heat maps do not contain the entire parts of objects.

Enhanced heat maps are then used to improve the performance of per-class saliency detection and object localization as well as semantic segmentation. We explain in detail how to apply improved heat maps to each task, and discuss the effectiveness of the proposed two-phase learning through various experiments.

In summary, this paper introduces the concept of two-phase learning for weakly supervised object localization. It allows the network to capture the full extent of the objects.  

\section{Related Works}
\label{sec:related_work}
In this section, we review previous studies that have sought to capture the spatial extent or the whole part of objects, not just the location of the most important part. Their goal coincides with that of two-phase learning. These studies can be broadly categorized into two types of approaches.

First, a group of approaches modify score aggregation methods in order to achieve a balance between the two most popular global pooling strategies: global max pooling (GMP)~\cite{Oquab15} and global average pooling (GAP)~\cite{zhou2016cvpr}. Since each of these pooling methods tends to underestimate or overestimate the extent of objects, respectively, finding a generalized model between these two extremes is essential. 
Pinheiro and Collobert~\cite{Pinheiro2015CVPR} aggregate activations into image-level scores through the log-sum-exp (LSE) pooling layer. In particular, Sun and Paluri~\cite{SunPCNB16pro} provide a comparison of GMP, GAP, and LSE pooling methods by showing the classification and localization performance of each method. Also, global weighted ranking pooling (GWRP) is proposed by Kolesnikov and Lampert ~\cite{kolesnikov2016seed} to properly combine properties of GMP and GAP. However, these methods are based on a user-parameter about the object size, which predetermines the portion of an image to be focused on.

The second group of methods employ external algorithms to obtain saliency masks or object proposals. Wei \textit{et al.}~\cite{Wei15stc} construct a new dataset consisting of images with a well-centered single object, and then apply the state-of-the-art saliency detection method proposed by Jiang \textit{et al.}~\cite{JiangWYWZL13drfi} to generate foreground/background masks. Qi \textit{el al.}~\cite{QiLSZJ16aug}, Pinheiro \textit{el al.}~\cite{Pinheiro2015CVPR}, and Bearman \textit{et al.}~\cite{BearmanRFL16wtp} make use of external region proposal methods to boost their performance. Selective search \cite{UijlingsSGS13ss}, CMPC~\cite{carreira2012pami}, BING \cite{ChengZLT14bing}, Objectness \cite{AlexeDF12obj}, and MCG \cite{ArbelaezPBMM14mcg} are the popular helpers. One approach with no such dependencies is suggested by  Saleh \textit{et al.}~\cite{SalehASPGA16builtin}. They extract saliency masks from the network itself by fusing feature maps from conv4 and conv5 layers. However, the aforementioned problem of a typical FCN is still inherent, and thus human annotation is further involved to achieve higher performance.

Our proposed method is fundamentally different from the previous approaches. We do not focus on determining aggregation methods but on finding more comprehensive features of objects. Thus, we are able to train the network to collect class-related regions without prior knowledge about the object size. Also, our approach relies on no external module that requires lower-than-image-level annotations.


\section{Two-Phase Learning}
\label{sec:two_phase_learning}
This section describes the dataset and the baseline network architecture used in our approach. We then go into detail about two-phase learning, which consists of the first phase learning, inference conditional feedback, and the second phase learning. 
We refer to each of the networks trained in the first and second phase learning as the first and the second network, respectively. Finally, we introduce the inference step where the two sets of heat maps obtained from both networks are combined.

\begin{figure*}[t]
\centering
\includegraphics[width=0.95\textwidth]{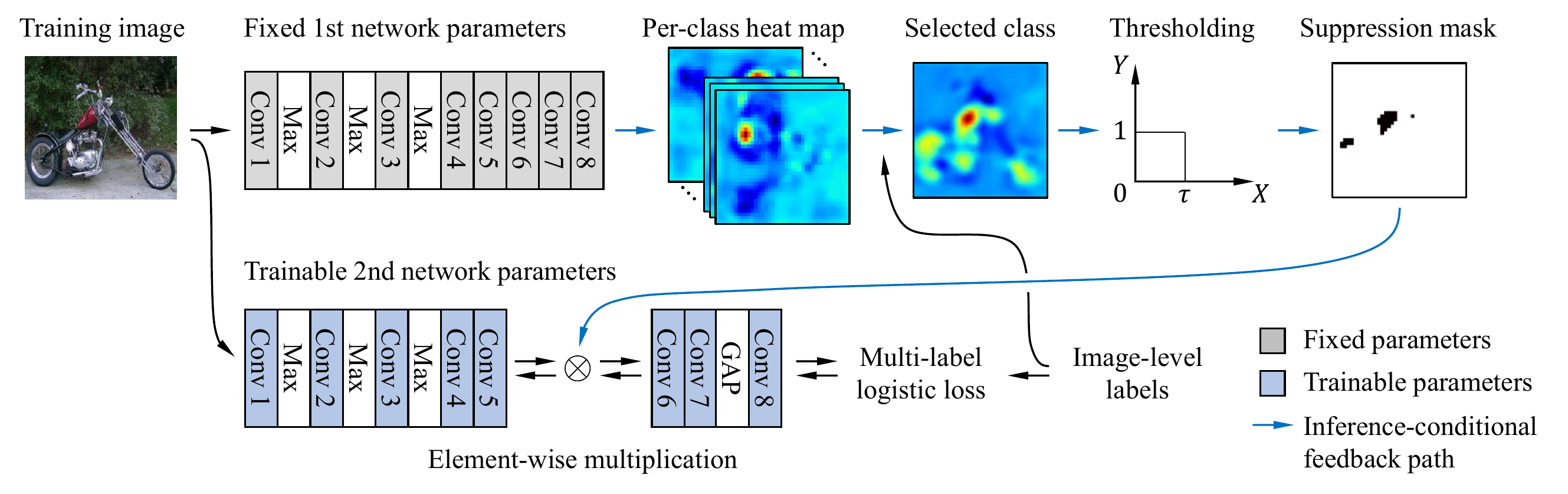}
\caption{The second phase learning. The overall process of inference conditional feedback is marked as blue arrows: The first network (with fixed layers, colored in gray) takes an input image and outputs heat maps. Only the heat maps corresponding to the classes present in the image labels are selected, and become a suppression mask after applying thresholding. The suppression mask is then element-wise multiplied with the conv5-3 output of the second network (with trainable layers, colored in blue). The forward and backward passes are marked as black arrows.}
\label{fig:feedback}
\end{figure*}
\vspace{-0.1in}

\subsection{Dataset}
We train on the Pascal VOC 2012 datasets. In practice, we use \textit{trainaug} part with 10,582 weakly annotated images of Pascal VOC 2012, as configured by \cite{HariharanABMM11}. The input images are rescaled to 321 $\times$ 321, as in \cite{kolesnikov2016seed}.

\subsection{Baseline Architecture}
\label{sec:Architecture}
The dominant paradigm of weakly supervised learning for object localization is to use a FCN with global pooling. The network is trained only by image-level supervision and generates heat maps for each class at the last convolutional layer. The global pooling layer then aggregates the heat maps for each class to compare with image-level labels. 

Among a number of weakly supervised FCNs, we build on a particular FCN proposed in \cite{zhou2016cvpr}. It is basically a modified VGG~\cite{Simonyan14c} variant, where fc6 and fc7 are converted to conv6 and conv7 and randomly initialized. GAP and a 20-way fully-connected layer are followed, and also pool4 and pool5 are removed.

This network has been imported as a component into one of the state-of-the-art techniques for weakly supervised semantic segmentation. Therefore, it is convenient to manifest the effect of our contribution by simply replacing the component with ours, and testing the segmentation performance of the entire system. Note that, however, our approach is not especially dependent on this very architecture, but can be applied to any types of existing FCNs.

\subsection{First Phase Learning}
In the first phase, a FCN is trained with a multi-label logistic loss for 20 foreground classes. The network is optimized via stochastic gradient descent (SGD) for 8,000 iterations, with a batch size of 15 and a weight decay of 0.0005. The learning rate is initially set to 0.001 and is reduced by a factor of 10 every 2000 iterations. At the inference stage, the network outputs class-specific heat maps; see \cite{zhou2016cvpr} for details.

\subsection{Inference Conditional Feedback}
The inference conditional feedback suppresses neurons not to fire repeatedly on the locations that had high activations in the first network. In order to realize this, we design a suppression mask to block the first highlighted regions during training. First, out of the 20 heat maps from the first network, we select only the heat maps that are relevant to a given image-level label. We then apply an inverse rectification: for each selected heat map, we apply hard thresholding by 60\% of the maximum value. In practice, we assign a value of zero to pixels above the threshold and one otherwise, as 
\begin{eqnarray}
   M_{supp,u}^{c} =  
\begin{cases}
    0,& \text{if }  \quad H_{u}^{c} > 0.6 \cdot max(H_{*}^{c})\\
    1,              & \text{otherwise}
\end{cases},
\label{eq:binary_sup_mask}
\end{eqnarray}
where $M_{supp,u}^{c} \in \mathbb{R} \textsuperscript{41 $\times$ 41}$,  $M_{u}^{c}\in \mathbb{R} \textsuperscript{41 $\times$ 41}$, $u$ and $c$ denote the binary suppression mask, per-class heat map, pixel position, and the indices of the classes present, respectively.

If there are multiple categories present, and consequently multiple binary suppression masks, they are combined by a logical \textit{AND} operation as
\begin{eqnarray}
   M_{supp,u} =  \prod_{c} M_{supp,u}^{c}.
\label{eq:logicalAND}
\end{eqnarray}
Finally, a resulting binary suppression mask $M_{supp,u}$ is fed to the second network to suppress neurons from being activated at the same locations as in the first network.

\subsection{Second Phase Learning}
\label{sec:Second_phase}
During the second phase, the first network with its fixed parameters is considered as a \textit{function} that takes an image as input and produces a binary suppression mask as output. \figref{fig:feedback} illustrates how this suppression mask is fed back into the training of the second network. Here, the second network has the same architecture as the first network.

As noted in \cite{GirshickDDM14}, all the layers up to the conv5-3 layer are regarded as feature extractors, where they learn the class-tuned representations. 
Based on the insight, we believe that it is semantically most appropriate to apply the feedback just after the conv5-3 layer of the second network. In practice, a suppression mask is multiplied with each channel of the conv5-3 output, element-wise. 
Thus, the forward pass with suppression mask is given as
\begin{eqnarray}
   {C^{'}}_{u}^{k} = M_{supp,u} \cdot {C^{}}_{u}^{k} \quad \forall \quad{k},
   \label{eq:EltwiseForward}
\end{eqnarray}
where ${C^{}}_{u}^{k}\in \mathbb{R} \textsuperscript{41 $\times$ 41}$ and ${C^{'}}_{u}^{k}\in \mathbb{R} \textsuperscript{41 $\times$ 41}$ denote activations before and after applying the suppression to the conv5-3 output, and $k$ denotes each channel of the conv5-3 output. Similarly, backward pass is given as
\begin{eqnarray}
   \frac{\partial L}{\partial {C^{}}_{u}^{k}} = M_{supp,u} \cdot \frac{\partial L}{\partial {C^{'}}_{u}^{k}} \quad \forall \quad{k},
   \label{eq:EltwiseBackward}
\end{eqnarray}
where $L$ denotes the output loss.
During the forward pass and backward update, the suppressed pixels are ignored. In other words, from the conv5-3 layer, activations on the previously important regions are dropped out by the feedback during the second phase.

The second network is subsequently trained to do image-level classification without the feature information that was most discriminative in the first phase. In this manner, the second network focuses on new features that can still be used to distinguish categories, and thus reveals more regions that were not highlighted in the first phase.

We can further think of the third or more phases using the next inference conditional feedback by lowering the threshold. However, as shown in \tabfref{tab:location}, the localization performance gradually decreases as the phase proceeds (the threshold of 40\% is used for the third phase). Therefore, only two phases of learning are considered throughout the applications of our approach.

\subsection{Inference} 
At the inference stage, the feedback is not defined. The first and second networks produce two sets of heat maps each in a single forward pass. The implementations on how to combine the two sets of heat maps will vary depending on the applications, as we will explain in \secref{sec:seg}, \secref{sec:saliency}, and \secref{sec:location}.

\begin{table*}[]
\centering
\resizebox{\textwidth}{!}{
\begin{tabular}{@{}c|ccccccccccccccccccccc|c@{}}
\hline
Method          & bg   & plane & bike & bird & boat & bottle & bus  & car  & cat  & chair & cow  & table & dog  & horse & motor & person & plant & sheep & sofa & train & tv   & mIoU \\ 
\hline  \hline
\textbf{Semi supervised:}& & & & & & & & & & & & & & & & & & & & & & \\
MIL+seg \cite{Pinheiro2015CVPR}& 79.6& 50.2& 21.6& 41.6& 34.9& 40.5& 45.9& 51.5& 60.6& 12.6& 51.2& 11.6& 56.8& 52.9& 44.8& 42.7& 31.2& 55.4& 21.5& 38.8& 36.9& 42.0 \\
MIL+bbox \cite{Pinheiro2015CVPR}& 78.6& 46.9& 18.6& 27.9& 30.7& 38.4& 44.0& 49.6& 49.8& 11.6& 44.7& 14.6& 50.4& 44.7& 40.8& 38.5& 26.0& 45.0& 20.5& 36.9& 34.8 &37.8\\
STC \cite{Wei15stc}& {84.5}& {68.0} &19.5 &60.5& {42.5}& {44.8}& {68.4}& {64.0}& 64.8& 14.5& 52.0 &22.8 &58.0 &55.3 &57.8 & {60.5}& {40.6}& 56.7& 23.0& {57.1}& 31.2& 49.8 \\
CheckMask \cite{SalehASPGA16builtin}& 86.4& 70.1& 21.7& 53.1& 52.5& 50.7& 70.9& 66.6& 63.2& 16.9& 45.8& 39.1& 61.1& 50.0& 56.8& 56.2& 40.0& 51.9& 29.3& 63.1& 35.9& 51.5 \\
 \hline \hline
 \textbf{Weakly supervised:}& & & & & & & & & & & & & & & & & & & & & & \\
EM-Adapt \cite{Papandreou_2015_ICCV}& 67.2& 29.2& 17.6& 28.6& 22.2& 29.6& 47.0& 44.0& 44.2& 14.6& 35.1& 24.9& 41.0& 34.8& 41.6& 32.1& 24.8& 37.4& 24.0& 38.1& 31.6&33.8\\ 
CCNN \cite{pathakICCV15ccnn} &68.5& 25.5& 18.0& 25.4 &20.2 &36.3 &46.8 &47.1& 48.0& 15.8& 37.9 &21.0& 44.5& 34.5 &46.2& 40.7& 30.4 &36.3& 22.2& 38.8 &36.9& 35.3\\
MIL+sppxl \cite{Pinheiro2015CVPR}& 77.2& 37.3 &18.4 &25.4& 28.2 &31.9 &41.6& 48.1& 50.7& 12.7 &45.7 &14.6 &50.9 &44.1&39.2 &37.9 &28.3& 44.0& 19.6 &37.6& 35.0& 36.6 \\
CheckMask-tags \cite{SalehASPGA16builtin}& 79.2 &60.1& 20.4 &50.7& \textbf{41.2} &\textbf{46.3} &62.6& 49.2& 62.3 &13.3& 49.7 &\textbf{38.1} &58.4& 49.0 &57.0& 48.2 &27.8& 55.1& 29.6& \textbf{54.6}& 26.6 &46.6\\
SEC (baseline) \cite{kolesnikov2016seed}& 82.4 & \textbf{62.9}  & \textbf{26.4} & 61.6 & 27.6 & 38.1   & 66.6 & 62.7 & {75.2} & \textbf{22.1}  & {53.5} & 28.3  & {65.8} &{57.8}  & \textbf{62.5}  & 52.5   & 32.5  & {62.6}  & {32.1} & 45.4  & {45.3} & {50.7} \\

Ours & \textbf{82.8} & 62.2 & {23.1}& \textbf{65.8}& 21.1& 43.1& \textbf{71.1}& \textbf{66.2}& \textbf{76.1}& {21.3}& \textbf{59.6}& {35.1}& \textbf{70.2}& \textbf{58.8}& {62.3}& \textbf{66.1}& \textbf{35.8}& \textbf{69.9}& \textbf{33.4}& 45.9&
\textbf{45.6}& \textbf{53.1}\\ 
\hline

\end{tabular}
}
\caption{Comparison of weakly supervised semantic segmentation methods on VOC 2012 ~\textit{segmentation, val.} set.}

\label{tab:segmentation}

\end{table*}

\begin{table*}[]
\centering
\resizebox{\textwidth}{!}{%
\begin{tabular}{@{}c|ccccccccccccccccccccc|c@{}}
\hline
Method          & bg   & plane & bike & bird & boat & bottle & bus  & car  & cat  & chair & cow  & table & dog  & horse & motor & person & plant & sheep & sofa & train & tv   & mIoU \\ 
\hline  \hline
\textbf{Semi supervised:}& & & & & & & & & & & & & & & & & & & & & & \\
MIL+seg \cite{Pinheiro2015CVPR}&78.7 &48.0 &21.2 &31.1 &28.4 &35.1 &51.4 &55.5 &52.8 &7.8 &56.2 &19.9 &53.8 &50.3 &40.0 &38.6 &27.8 &51.8 &24.7 &33.3 &46.3 &40.6\\
MIL+bbox \cite{Pinheiro2015CVPR}&76.2 &42.8 &20.9 &29.6 &25.9 &38.5 &40.6 &51.7 &49.0 &9.1 &43.5 &16.2 &50.1 &46.0 &35.8 &38.0 &22.1 &44.5 &22.4 &30.8 &43.0 &37.0 \\
STC \cite{Wei15stc}&85.2 &62.7 &21.1 &58.0 &31.4&55.0 &68.8 &63.9 &63.7 &14.2 &57.6 &28.3 &63.0 &59.8 &67.6 &61.7 &42.9 &61.0 &23.2 &52.4 &33.1 &51.2\\
CheckMask \cite{SalehASPGA16builtin}&87.4 &65.7 &26.0&64.2&43.7&53.2&72.6&63.6&59.5&17.1&48.0&43.7&61.2&52.0&69.3&54.8&43.0 &50.3 &34.6 &59.2 &42.0 &52.9 \\

 \hline \hline
 \textbf{Weakly supervised:}& & & & & & & & & & & & & & & & & & & & & & \\
EM-Adapt \cite{Papandreou_2015_ICCV}&76.3 &37.1 &21.9 &41.6 &26.1 &38.5 &50.8 &44.9 &48.9 &16.7 &40.8 &29.4 &47.1 &45.8 &54.8 &28.2 &30.0 &44.0 &29.2 &34.3 &46.0 &39.6 \\ 
CCNN \cite{pathakICCV15ccnn}&- &24.2 &19.9 &26.3 &18.6 &38.1 &51.7 &42.9 &48.2 &15.6 &37.2 &18.3 &43.0 &38.2 &52.2 &40.0 &33.8 &36.0 &21.6 &33.4 &38.3 &35.6\\
MIL+sppxl \cite{Pinheiro2015CVPR}& 74.7&38.8 &19.8 &27.5 &21.7 &32.8 &40.0 &50.1 &47.1 &7.2 &44.8 &15.8 &49.4 &47.3 &36.6 &36.4 &24.3 &44.5 &21.0 &31.5 &41.3 &35.8\\
CheckMask-tags \cite{SalehASPGA16builtin}& 80.3 &57.5 &24.1 &66.9 &\textbf{31.7} &43.0 &67.5 &48.6 &56.7 &12.6 &50.9 &\textbf{42.6} &59.4 &52.9 &65.0 &44.8 &\textbf{41.3} &51.1 &33.7 &\textbf{44.4} &33.2 &48.0\\

SEC (baseline) \cite{kolesnikov2016seed}& \textbf{83.5} & 56.4  & \textbf{28.5} & 64.1& 23.6 & 46.5 & \textbf{70.6} & 58.5& 71.3 & \textbf{23.2}  & {54.0} & 28.0  & {68.1} &\textbf{62.1}  & {70.0}  & 55.0   & 38.4  & {58.0}  & {39.9} & 38.4  & \textbf{48.3} & {51.7} \\

Ours & 83.4 & \textbf{62.2} & {26.4}& \textbf{71.8}& 18.2& \textbf{49.5}& 
66.5& \textbf{63.8}& \textbf{73.4}& {19.0}& \textbf{56.6}& {35.7}& \textbf{69.3}& {61.3}& \textbf{71.7}& \textbf{69.2}& {39.1}& \textbf{66.3}& \textbf{44.8}& 35.9&
45.5& \textbf{53.8}\\ 
\hline

\end{tabular}
}\caption{Comparison of weakly supervised semantic segmentation methods on VOC 2012 \textit{segmentation, test.} set.}
\label{tab:segmentation}

\end{table*}


\section{Semantic Segmentation Experiments}
\label{sec:seg}
In the task of semantic segmentation, each pixel in the image is classified into one of 21 categories including the background. However, in a weakly supervised setting, the network cannot explicitly learn the information about object boundaries or sizes. Therefore, to successfully perform this task, it is essential to initially retrieve accurate localization cues. Most techniques for weakly supervised segmentation internally train FCNs and obtain localization cues from the heat maps for each category. 

The heat maps obtained via two-phase learning can cover not only the most discriminative parts of objects but also the whole parts. Thus, the quality of our localization cues is enhanced, and the performance of semantic segmentation is also increased accordingly. In order to verify this, we apply our two-phase learning algorithm to the SEC model \cite{kolesnikov2016seed}, one of the state-of-the-art methods for weakly supervised semantic segmentation.

In this section, we briefly review our baseline segmentation network, SEC, and describe how the localization cues are complemented via two-phase learning. we then experiment on semantic segmentation using the localization cues. Finally, we report and analyze the results.


\subsection{Review of SEC Architecture}
\label{sec:SEC}

As introduced in~\cite{kolesnikov2016seed}, SEC stands for \textit{seed}, \textit{expand}, and \textit{constrain}. They are referred to as three important principles in weakly supervised semantic segmentation. First, a \textit{seed} is a module to provide localization cues to the main segmentation network. The segmentation network is implicitly supervised to match the retrieved localization cues. 
Next, \textit{expand} considers how to aggregate heat maps into image-level scores. It encourages the responses on promising locations to be high and to be consistent with image-level labels. As a new pooling strategy, global weighted rank pooling (GWRP) is proposed in order to recover the spatial information that will be lost in the aggregation process. Lastly, \textit{constrain} is a module that constrains the results of the segmentation networks to follow the boundaries of objects. In practice, fully-connected conditional random fields (dense CRF) \cite{ToyodaH08crf} are used.

\subsection{Two-phase Learning for Localization Cues}
A set of localization cues, \textit{seed}, is \textit{a cornerstone} for a segmentation network to build on. In the context of the SEC model \cite{kolesnikov2016seed}, the localization cues refer to a set of class-specific binary masks that are obtained by a thresholding operation: for each per-class heat map, all pixels with a score larger than 20\% of the maximum score are selected.

The localization cues obtained using heat maps from a conventional FCN are considered reliable only for the object positions, so they remain \textit{weak}, as noted in~\cite{Pinheiro2015CVPR,Bilen16,kolesnikov2016seed}. With our proposed two-phase learning, the heat maps become more comprehensive. As a result, the localization cues for semantic segmentation also become more \textit{powerful}.

In practice, we have two sets of heat maps from the first and second networks. In order to integrate the information on object regions, we merge the two heat maps via \textit{weighted map voting}, which will be described in detail in \secref{sec::merging}. For the background class, as in~\cite{kolesnikov2016seed},  we imported the network implementation proposed in~\cite{SimonyanVZ13bg}, which generates class-agnostic saliency detection based on the image gradient. The inferred localization cues are used to supervise semantic segmentation task.

\subsection{Merging Heat Maps from Both Phases}
\label{sec::merging}
To effectively combine two heat maps, we consider a simple post-processing technique, \textit{weighted map voting}. We assume that a per-class probability score given by the network represents how confident the network is about the heat map of the same class. That is, if the first network predicts a high probability for a specific class, the information in the corresponding heat map is more confident than that of the second network, which predicts a lower probability for the same class.

Following this insight, \textit{weighted map voting} is integrated into the system by multiplying the per-class heat maps ${H\textsuperscript{c}}$ by its class probability scores ${p\textsuperscript{c}}$. We then merge the resulting maps by taking the pixel-wise maximal values between the two multiplications, that is:  
\vspace{-1mm}
\begin{eqnarray}
   H_{u}^{c} =  max( p_{1st}^{c} * H_{1st,u}^{c}, p_{2nd}^{c} * H_{2nd,u}^{c}),
\label{eq:weightedMax}
\end{eqnarray}
where the subscripts $u$ and $1st$ and $2nd$ denote the pixel position and the first and second networks, respectively.

\subsection{Improving Segmentation Network}

Our baseline segmentation network, SEC \cite{kolesnikov2016seed}, performs best when trained with all three losses of \textit{seed}, \textit{expand}, and \textit{constrain}. In its original form, it achieves an average intersection-over-union scores of 50.7\%, which is 0.3\% higher than the same network trained with only \textit{seed} and \textit{constrain} losses. 

Since our two-phase learning enables the localization cues to cover wider object regions in addition to the first predicted locations, it provides the segmentation network with richer information for object localization. In other words, our heat maps are able to perform both \textit{seeding} and \textit{expanding} roles in their former sense. Therefore, we use the only \textit{seed} and \textit{constrain} loss terms to train the segmentation network whose localizing module is replaced by our improved method. At inference, the predicted segmentation masks are rescaled to the size of their original images and refined by dense CRF \cite{ToyodaH08crf}.

\begin{figure*}
\begin{center}
\def\arraystretch{0.5}
\begin{tabular}{@{}c@{\hskip 0.01\linewidth}c@{\hskip 0.01\linewidth}c}
\includegraphics[width=0.33\linewidth]{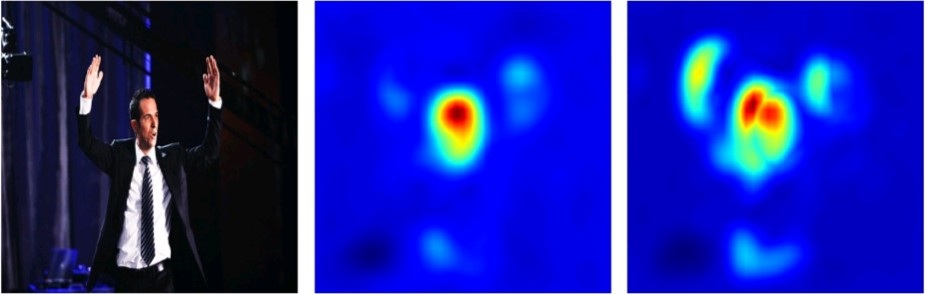} &
\includegraphics[width=0.33\linewidth]{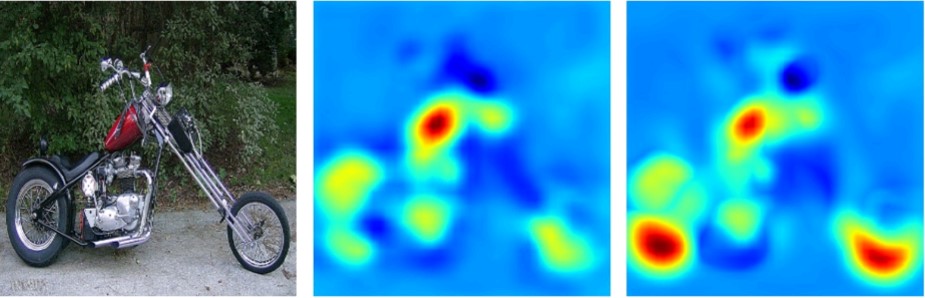} &
\includegraphics[width=0.33\linewidth]{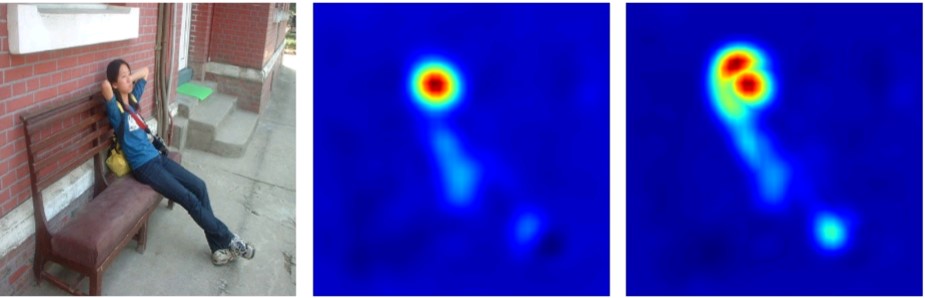} \\
\includegraphics[width=0.33\linewidth]{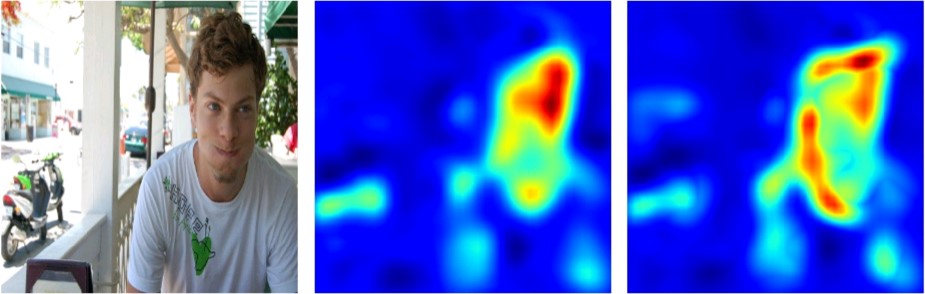} &
\includegraphics[width=0.33\linewidth]{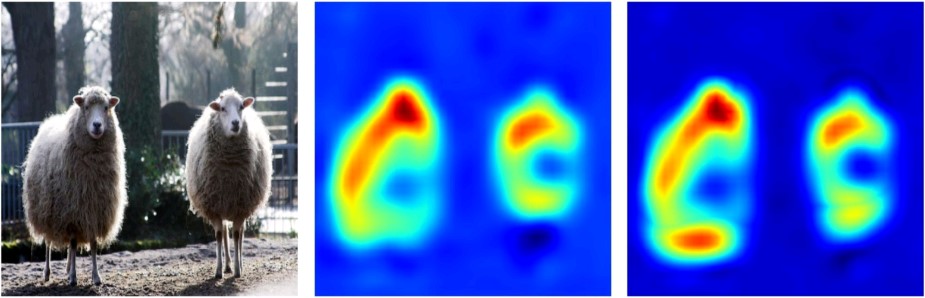} &
\includegraphics[width=0.33\linewidth]{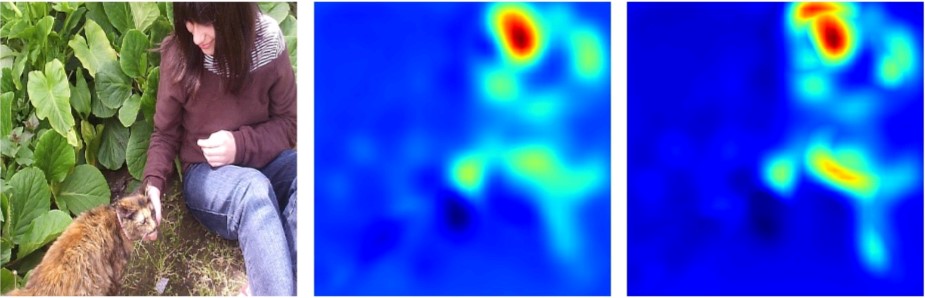} \\
\end{tabular}
\end{center}
\vspace{-0.1in}
\caption{Object saliency detections using the first network (column 2,5, and 8) and proposed method (column 3,6, and 9).}
\label{fig:saliency}
\end{figure*}

\subsection{Evaluation}
To evaluate the contribution of our two-phase learning on the semantic segmentation, we use the metric of intersection-over-union scores, following the protocol of Pascal VOC 2012 semantic segmentation challenge~\cite{Everingham10voc}. We evaluated the results on 1,449 images in the \textit{validation} part of the Pascal VOC 2012 segmentation dataset.

\subsection{Results and Discussion}
\tabfref{tab:segmentation} compares the numeric results of our approach with those of previous weakly supervised approaches. For reference, we also provide the results of other methods that utilize additional annotations. They require either additional data from \textit{Flickr} and an external saliency detector pretrained by pixel-level supervision~\cite{Wei15stc} or user clicks~\cite{SalehASPGA16builtin} or region proposals such as selective search and MCG~\cite{Pinheiro2015CVPR}. Since they are not trained with purely image-level annotations, we refer to them as semi-supervised learning. In this regard, only EM-Adapt~\cite{Papandreou_2015_ICCV}, CCNN~\cite{pathakICCV15ccnn}, MIL+sppxl~\cite{Pinheiro2015CVPR}, CheckMask-tags~\cite{SalehASPGA16builtin}, and SEC~\cite{kolesnikov2016seed} would be fair comparisons with ours. 
Among them, we achieved the best mIoU scores; 53.1\% on VOC-val and 53.8\% on VOC-test, which improve upon the SEC baseline by 2.4\% and 2.1\% for each set.

\begin{figure*}
\begin{center}
\def\arraystretch{0.5}
\begin{tabular}{@{}c@{\hskip 0.01\linewidth}c@{\hskip 0.01\linewidth}c}
\includegraphics[width=0.5\linewidth]{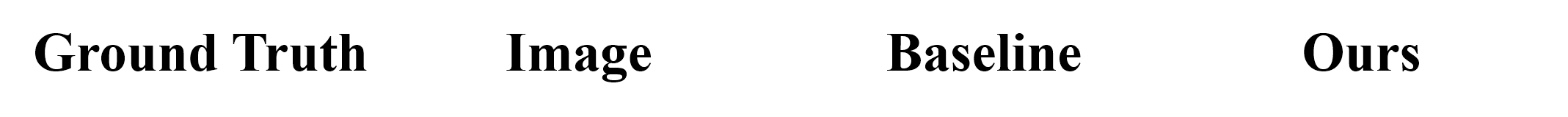} &&
\includegraphics[width=0.5\linewidth]{figures/segm/title3.png} \\
\includegraphics[width=0.5\linewidth]{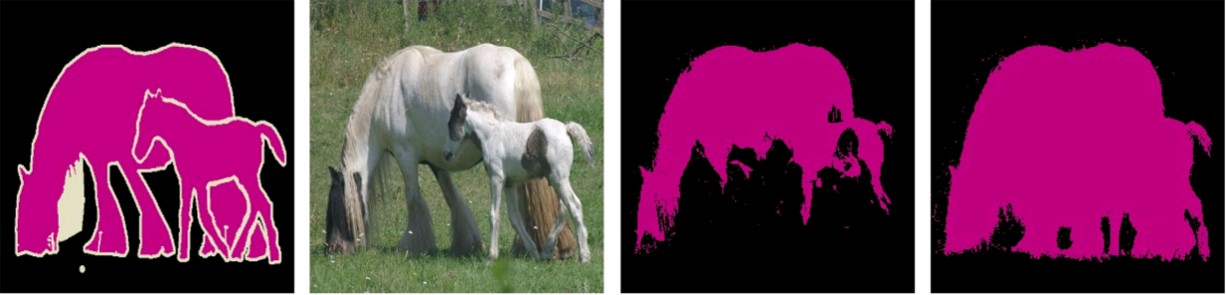} &&
\includegraphics[width=0.5\linewidth]{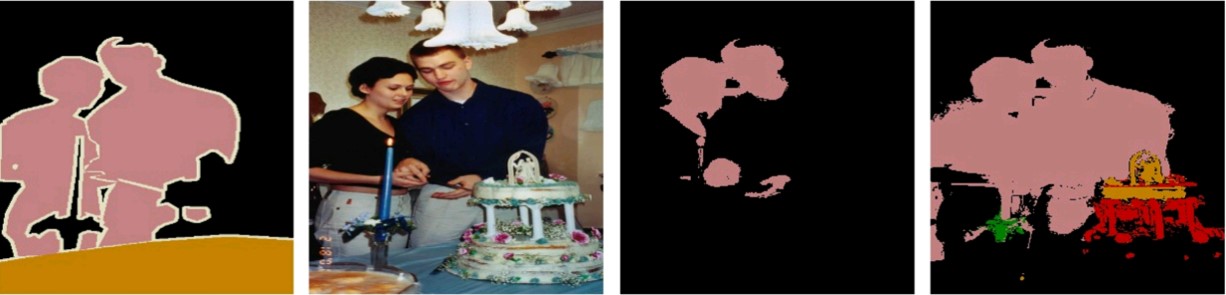} \\
\includegraphics[width=0.5\linewidth]{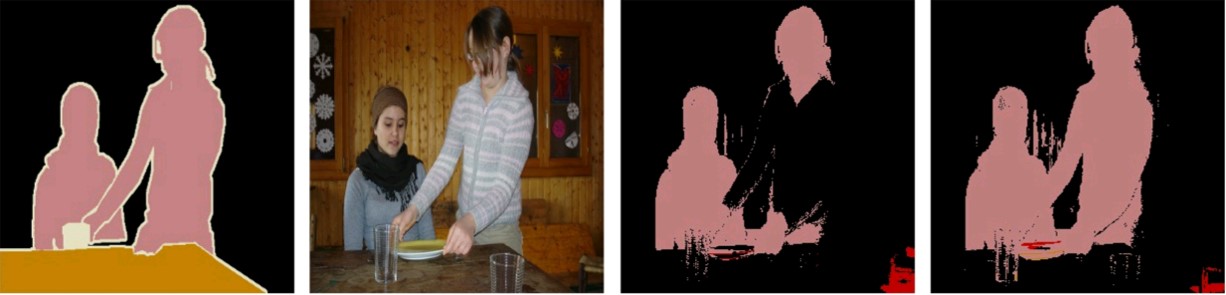} &&
\includegraphics[width=0.5\linewidth]{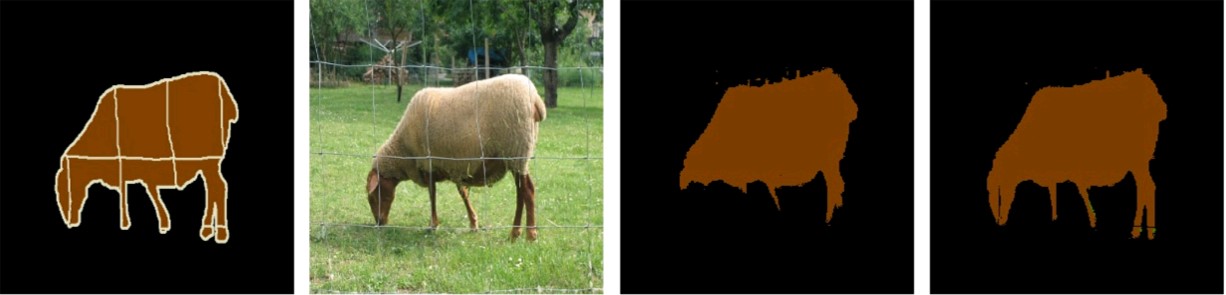} \\
\includegraphics[width=0.5\linewidth]{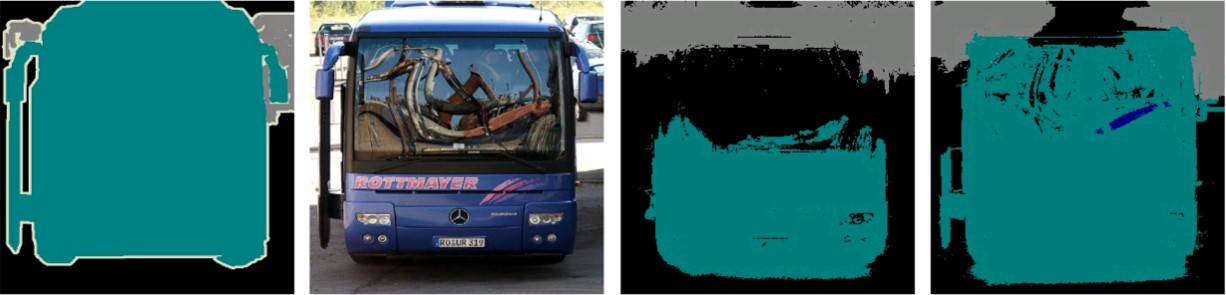} &&
\includegraphics[width=0.5\linewidth]{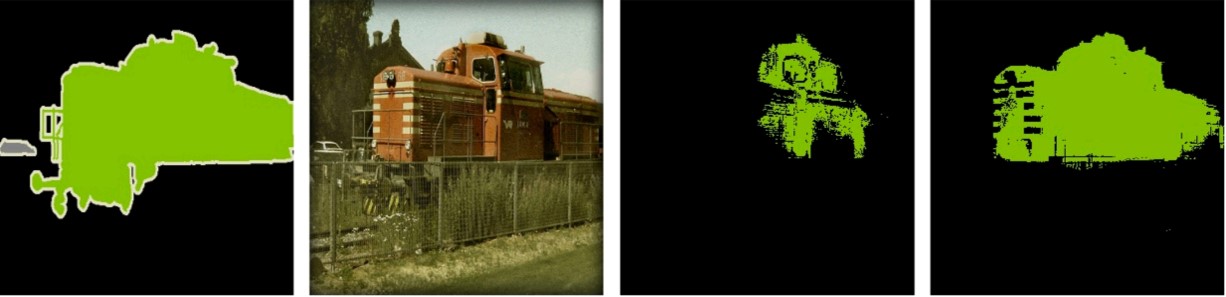} \\
\includegraphics[width=0.5\linewidth]{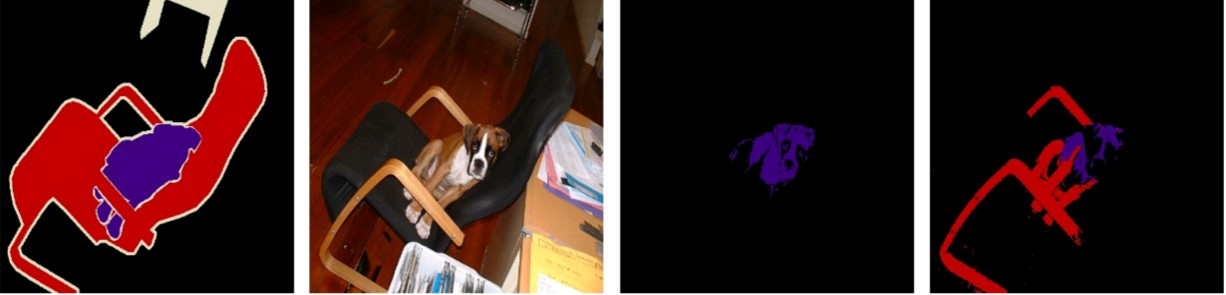} &&
\includegraphics[width=0.5\linewidth]{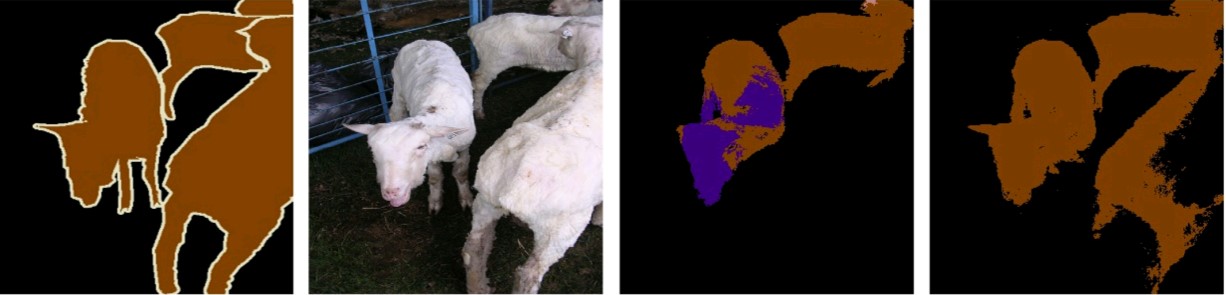} \\
\includegraphics[width=0.5\linewidth]{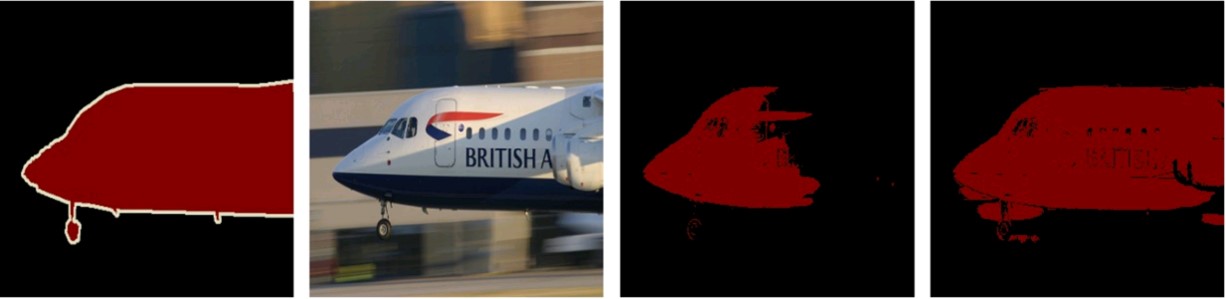} &&
\includegraphics[width=0.5\linewidth]{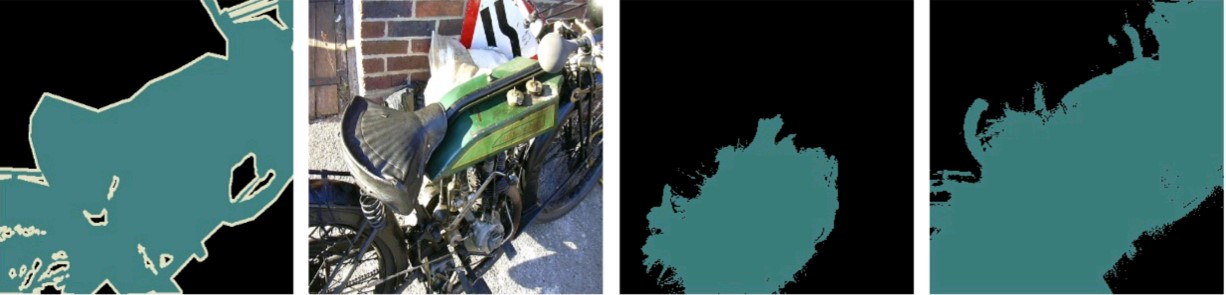} \\
\includegraphics[width=0.5\linewidth]{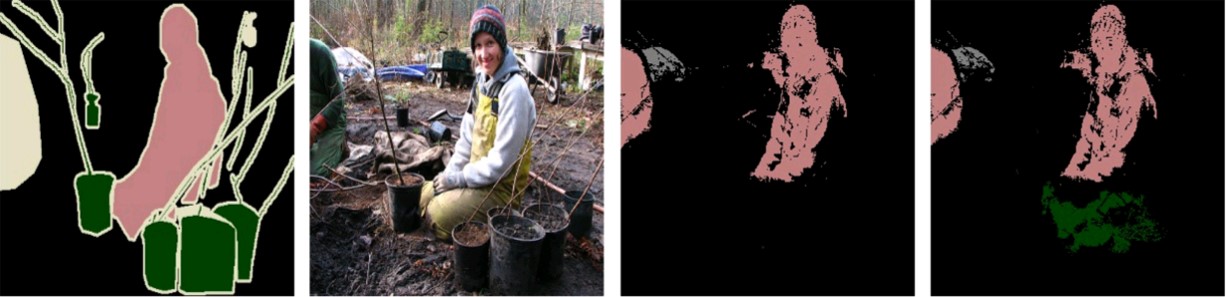} &&
\includegraphics[width=0.5\linewidth]{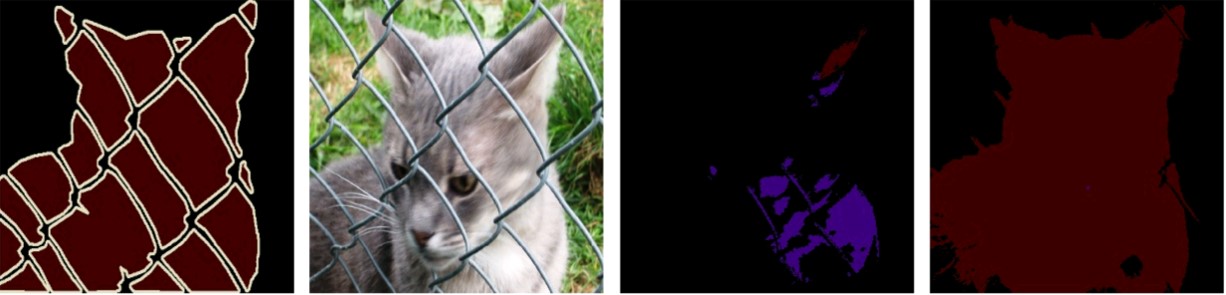} \\
\includegraphics[width=0.5\linewidth]{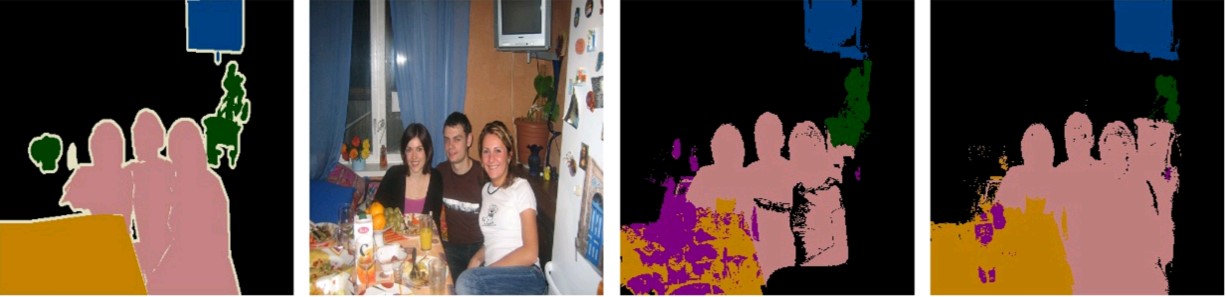} &&
\includegraphics[width=0.5\linewidth]{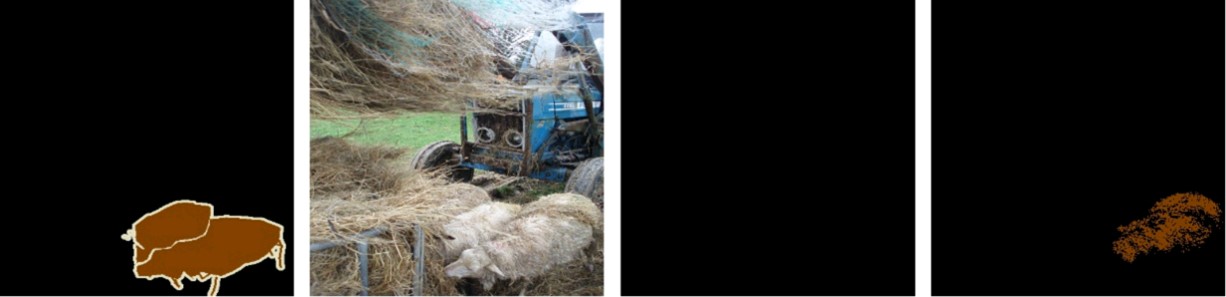} \\
\end{tabular}
\end{center}
\vspace{-0.2in}
\caption{Qualitative segmentation results. Discovering more object regions (on VOC 2012 \textit{segmentation, val.} set).}
\label{fig:segm1}
\end{figure*}

\begin{figure*}
\begin{center}
\def\arraystretch{0.5}
\begin{tabular}{@{}c@{\hskip 0.01\linewidth}c@{\hskip 0.01\linewidth}c}
\includegraphics[width=0.5\linewidth]{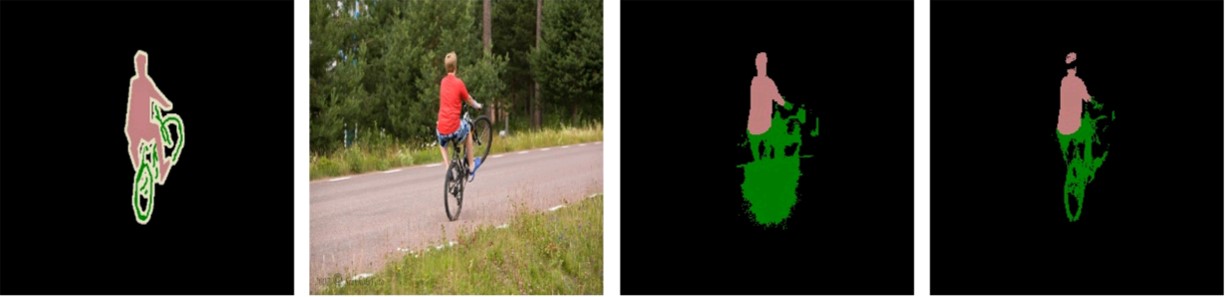} &&
\includegraphics[width=0.5\linewidth]{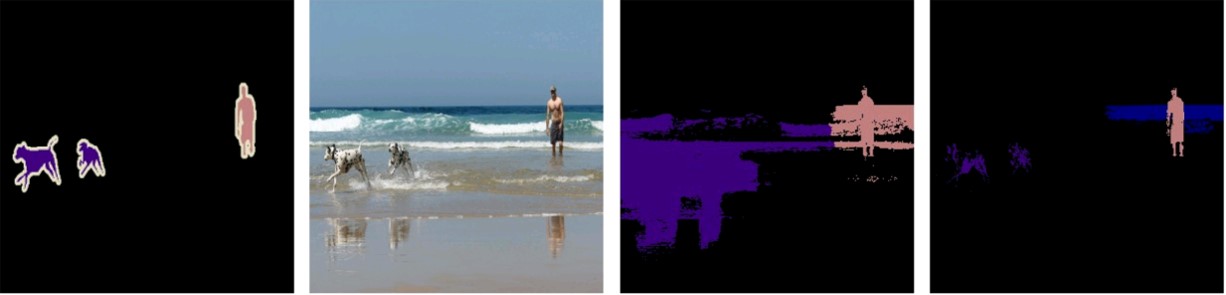} \\
\includegraphics[width=0.5\linewidth]{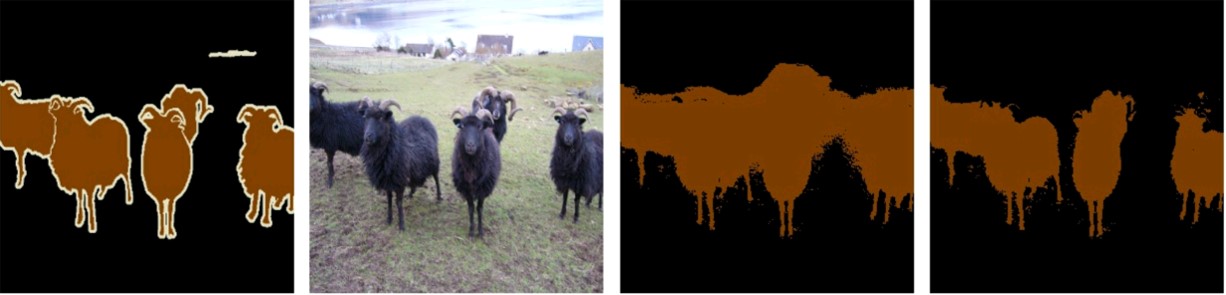} &&
\includegraphics[width=0.5\linewidth]{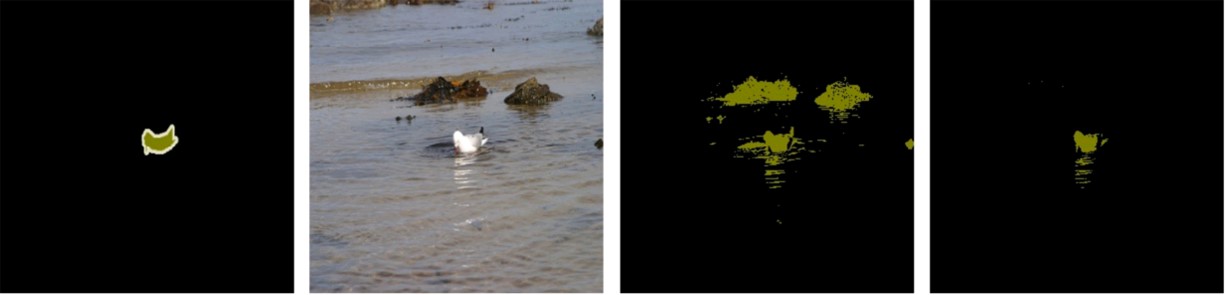} \\
\end{tabular}
\end{center}
\vspace{-0.2in}
\caption{Qualitative segmentation results. Expanding up to reasonable extent (on VOC 2012 \textit{segmentation, val.} set).}
\label{fig:segm2}
\vspace{-0.1in}
\end{figure*}

\begin{figure}
\begin{center}
\def\arraystretch{0.5}
\begin{tabular}{@{}c@{\hskip 0.01\linewidth}c@{\hskip 0.01\linewidth}c}
\includegraphics[width=0.99\linewidth]{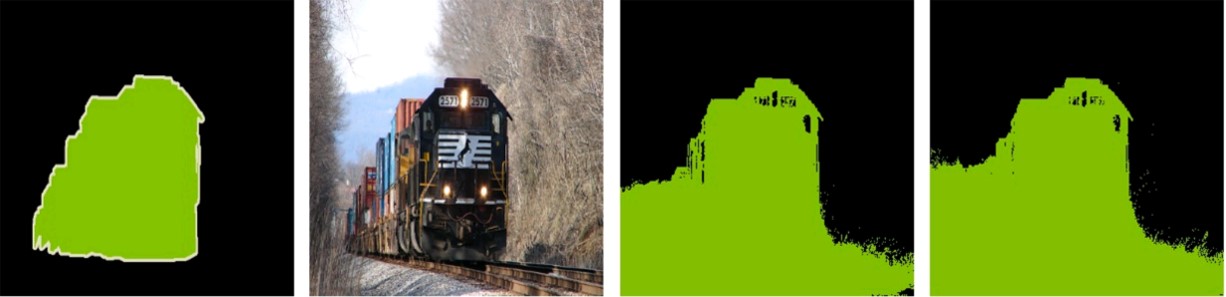} \\
\includegraphics[width=0.99\linewidth]{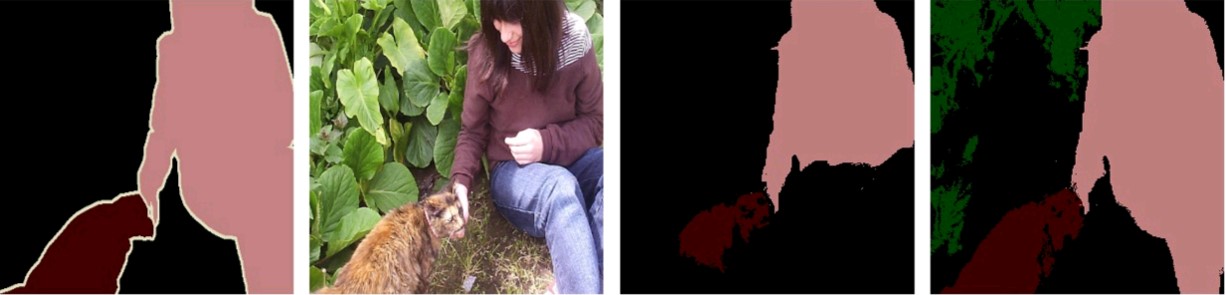} \\
\end{tabular}
\end{center}
\vspace{-0.2in}
\caption{Qualitative segmentation results. Some failure cases (on VOC 2012 \textit{segmentation, val.} set).}
\label{fig:segm3}
\vspace{-0.1in}
\end{figure}

\figref{fig:segm1}, \figref{fig:segm2}, and \figref{fig:segm3} illustrate visual comparisons of the segmentation predicted by the baseline~\cite{kolesnikov2016seed} and ours.

\paragraph{Discovering More Object Regions}\quad A chronic problem of weakly supervised semantic segmentation is that the segmentation covers only parts of objects. This is because their heat maps tend to focus only on the most discriminative parts, e.g. \textit{a person's face}. 
In particular, when objects are cropped or partially occluded, the object is often totally ignored in the prediction.
We observe that our two-phase learning is able to overcome this problem. On this level, \figref{fig:segm1} compares qualitative results of the baseline and ours. The segmentation network trained in our method covers more object regions than the baseline~\cite{kolesnikov2016seed}. More specifically, it either discovers other parts of objects, e.g. \textit{a torso, arms, and legs of a person}, or reveals new instances that have not been found before.

\paragraph{Expanding up to Reasonable Extent}\quad
As mentioned in \secref{sec:SEC}, various aggregation methods often fail to accurately estimate the extent of objects. This is because they enforce the network to \textit{expand} to a certain degree. This often causes unreasonable expansions, as shown in \figref{fig:segm2}. However, our approach is immune to this problem. The reason is that our system determines what additional features should be considered important. Therefore, the combination of the heat maps from the first and second networks does not simply widens the segmentation but also restricts it to fall inside the class-related regions. This method of propagation allows our approach to successfully remove the unreasonable expansions that happened in the baseline segmentation.

\paragraph{Failure Cases}\quad
Like typical weakly supervised segmentation techniques, our segmentation also has a problem distinguishing objects that co-occur almost always, e.g. \textit{trains vs. tracks}, as shown in \figref{fig:segm3}. Another failure case arises rarely, when the newly found regions do not belong to the the predicted class, e.g. \textit{plants but not potted}. We believe this is because the newly highlighted features in the second phase are sometimes not discriminative enough to exclude such confusing regions. This implies that the two-phase learning will have an upper bound on the degree to which the important parts are suppressed, as noted in \secref{sec:Second_phase}.

\paragraph{Scope}\quad
In order to demonstrate that our method can be applied to other semantic segmentation methods using heat maps, we applied our method to CCNN~\cite{pathakICCV15ccnn}, and confirmed that the benefits of our method are consistent: Our approach achieves an mIoU score of 35.7\% on VOC-val, outperforming the CCNN baseline which achieves 34.5\% (what we could reproduce) by highlighting the second most important parts that are not found in the baseline. This implies that our two-phase learning is not limited to either the SEC model or the CAM~\cite{zhou2016cvpr} module, but is more generally applicable to other segmentation systems.

\section{Per-Class Saliency Prediction Experiments}

\label{sec:saliency}
In this section, we demonstrate that the two sets of heat maps obtained via two-phase learning can synergize each other to capture the complete object. Here, we consider the heat maps as per-class saliency maps. Accordingly, we investigate whether those saliency maps are consistent with the ground truth segmentation masks. The two sets of heat maps are combined via \textit{weighted map voting}, as given in \eqnref{eq:weightedMax}.

\subsection{Evaluation}

In order to evaluate the quality of our heat maps, we only consider the heat maps whose corresponding class is present in the images. Similarly, we extract per-class saliency masks only for the classes present, from the ground truth segmentation. We use these as our ground truth saliency masks. In practice, 2,148 pairs of a per-class heat map and the ground truth saliency mask are collected for 1,440 images in Pascal VOC 2012 \textit{val.} set. Each pixel in those heat maps has a response value that we consider as a confidence value, and we generate a precision-recall curve and compute the average precision (AP).

\subsection{Results}

A set of our heat maps combined via \textit{weighted map voting} achieves an AP of 37.7\%, which is 5.5\% higher than that (32,5\%) achieved using only the first heat maps. \figref{fig:saliency} illustrates the qualitative results: in our combined heat maps, the regions highlighted by both networks are revealed on object-relevant locations, e.g. \textit{the hands of a person, wheels of a motorcycle}, and \textit{a person's feet}. 

\begin{table*}[]
\centering
\resizebox{\textwidth}{!}{%
\begin{tabular}{@{}c|cccccccccccccccccccc|c@{}}
\hline
Phase           & plane & bike & bird & boat & bottle & bus  & car  & cat  & chair & cow  & table & dog  & horse & motor & person & plant & sheep & sofa & train & tv   & mAP  \\ 
\hline  \hline
Center & 86.0 & 56.6 & 64.8 & 41.6 & 18.0 & 82.5 & 30.0 & 87.5 & 23.3 & 73.9 & 24.5 & 75.3 & 83.1 & 65.9 & 54.2 & 17.6 & 66.1 & 52.1 & 78.4 & 30.3 & 55.6\\ \hline
First  & 98.7 & 94.4 & 93.2 & 88.5 & 67.2 & 93.6 & 81.3 & 99.0 & 65.0 & 94.5 & 67.4 & 96.7 & 98.8 & 95.9 & 92.6 & 72.0 & 98.5 & 88.8 & 92.1 & 83.8 & 88.1\\
Second & 98.1 & 89.9 & 92.8 & 75.1 & 52.7 & 90.8 & 76.7 & 97.2 & 56.4 & 95.9 & 38.8 & 97.4 & 98.7 & 95.1 & 91.2 & 69.9 & 97.5 & 78.1 & 82.7 & 77.6 & 82.6 \\
Third  &94.6 &89.3 &88.5 &38.0 & 32.8 &86.0 &65.2 &96.4 &31.7 &93.9 &24.8 &95.1 &93.2 &89.1 &71.2 &27.2 &92.1 &43.4 &92.3 &64.8 & 70.5\\
\hline
\end{tabular}
}\caption{Object location prediction for each phase on VOC 2012 \textit{main, val.} set.}
\label{tab:location}
\end{table*}
\vspace{-0.1in}

\begin{figure*}
\begin{center}
\def\arraystretch{0.5}
\begin{tabular}{@{}c@{\hskip 0.01\linewidth}c@{\hskip 0.01\linewidth}c@{\hskip 0.01\linewidth}c@{\hskip 0.01\linewidth}c@{}}
\includegraphics[width=0.17\linewidth]{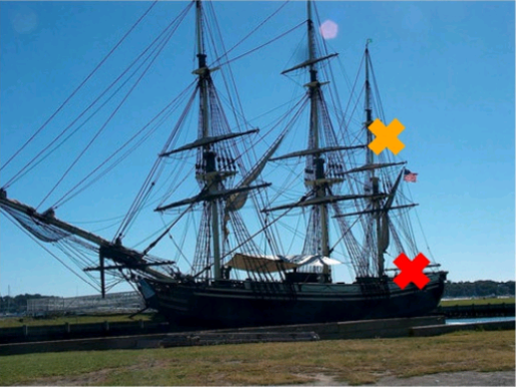} &
\includegraphics[width=0.17\linewidth]{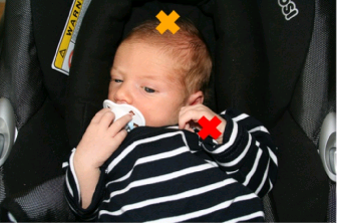} &
\includegraphics[width=0.17\linewidth]{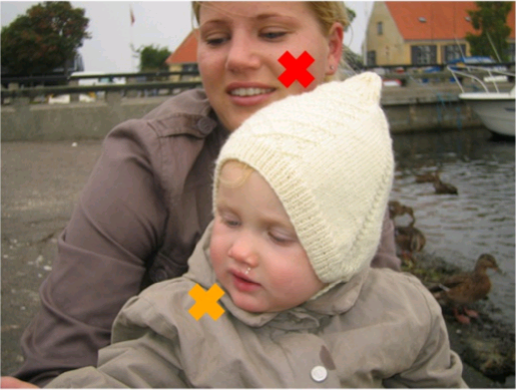} &
\includegraphics[width=0.17\linewidth]{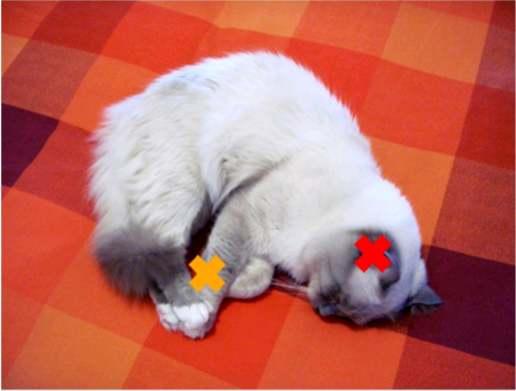} &
\includegraphics[width=0.17\linewidth]{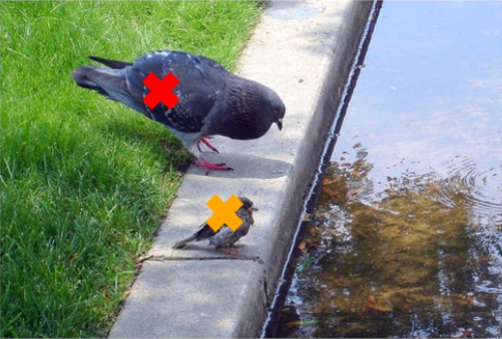} \\
\includegraphics[width=0.17\linewidth]{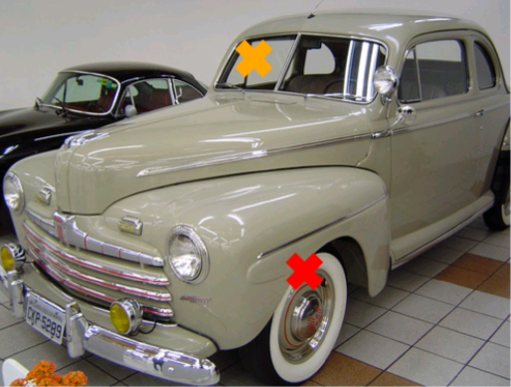} &
\includegraphics[width=0.17\linewidth]{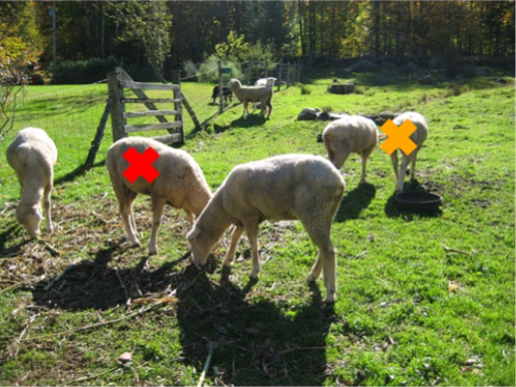} &
\includegraphics[width=0.17\linewidth]{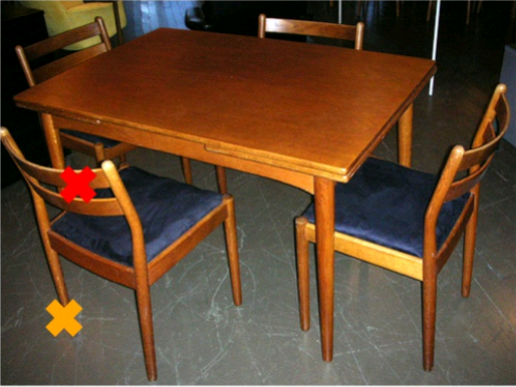} &
\includegraphics[width=0.17\linewidth]{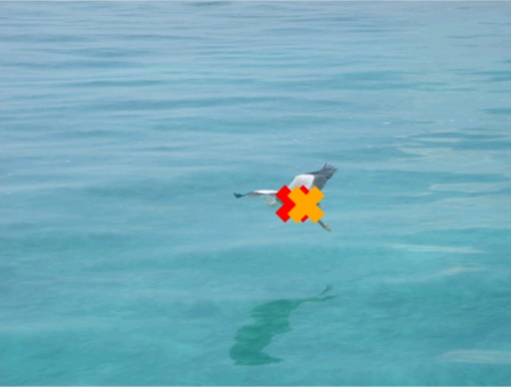} &
\includegraphics[width=0.17\linewidth]{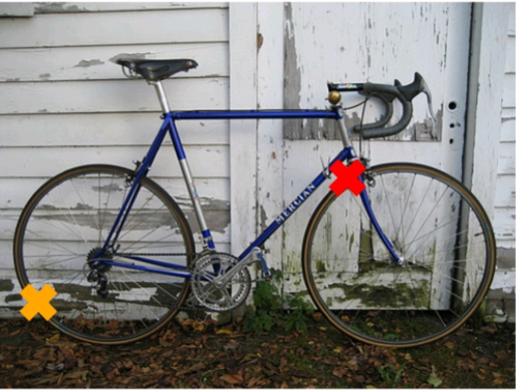} \\
\end{tabular}
\end{center}
\vspace{-0.1in}
\caption{Object location predictions of the first (red) and second (orange) networks.}
\label{fig:location}
\end{figure*}


\section{Location Prediction Experiments}
\label{sec:location}
The proposed two-phase learning allows the second network to focus on the valuable features that have not been discovered in the first phase learning. In the previous section, we have shown that those newly revealed features can be combined with the first features to better capture the extent of objects. However, in this section, we also demonstrate that the different features highlighted by each of the first and second networks are semantically consistent with the distinctive parts of objects.

Here, we experiment on 5,823 images and the ground truth bounding boxes of the Pascal VOC 2012 main \textit{val.} set.


\subsection{Evaluation}
In order to pinpoint the locations which the networks focus on, we consider the pixel of the maximal response of a per-class heat map as the predicted object location. For quantitative evaluation, we use the criteria introduced in \cite{Oquab15}. First, the heat maps are rescaled to their original image size using bilinear interpolation. With 18-pixel tolerance, the predicted location within any ground truth bounding boxes of the target category is counted as correct and false negative otherwise, see \cite{Oquab15} for details. For each image, for each class, the maximal response is considered as the confidence for the prediction, and this is then used to compute AP. Note that the heat maps from each network are not combined here but investigated separately because only the maximal value locations are considered.

Moreover, in order to confirm that the features that are considered important in both networks do not overlap, we measured the Euclidean pixel distance between the predicted locations of the first and second networks.

\subsection{Results and Discussion}

The per-class precisions of the location prediction for Pascal VOC are summarized in \tabfref{tab:location}. To show the difficulty of the location prediction task, we report the performance of our naive baseline, \textit{center}, which predicts the center of the image as the object location. 

As it has been widely noted in the literature \cite{zhou2016cvpr, Oquab15, SimonyanVZ13bg, kolesnikov2016seed} that weakly supervised FCNs reliably predict approximate positions of objects, our first network also successfully captures object locations, achieving an mAP of 88.1\%. However, our second network is at a great disadvantage in predicting object locations because the most discriminative parts of objects have not been shown during training. Nevertheless, the second network was able to highlight the next most important parts with a small performance reduction of 5.5\%, achieving an mAP of 82.6\%.  

Likewise, as shown in the previous experiments, the second network tends to highlight either different important parts of objects, e.g. \textit{sails of a boat, pillars of a car}, or other instances even of small sizes, e.g. \textit{a bird in front}. Also, even when the object region is small, it maintains the ability to predict the location, e.g. \textit{a small bird flying}, implying that the second learned features are also representative of the object. \figref{fig:location} visualizes some pairs of predictions. 

In most cases, two networks focus on different parts of images. The average Euclidean distance of the predictions of the two networks appeared to be 69 pixels. Considering that the average size of the images in the Pascal VOC 2012 dataset is 390 $\times$ 470, it is shown that the second network found fairly distant objects from those detected by the first network. Consequently, we demonstrate that different features highlighted in both networks can complement each other to localize objects.


\section{Conclusion}
Weakly supervised object localization has an inherent weakness that it often fails to capture the extent of objects because the network focuses only on the most distinctive parts of the objects. In this paper, we propose a two-phase learning algorithm that can fundamentally mitigate this problem. We have been motivated by the insight that if we retrain the network while covering the most discriminative parts of the objects, it will highlight feature regions that are different from the first, while those features still fall inside the range of the objects. We propose inference conditional feedback in order to train an additional network in this manner. Finally, the heat maps of the first and second networks are combined to enhance object localization. Experiments on semantic segmentation, object saliency detection, and object location prediction tasks have shown the effectiveness of our two-phase learning on the challenging Pascal VOC 2012 dataset.

\paragraph{Acknowledgements}
This work was supported by DMC R\&D Center of Samsung Electronics Co.



\clearpage
\clearpage

{\small
\bibliographystyle{ieee}
\bibliography{egbib}
}

\end{document}